\documentclass[runningheads, orivec]{llncs}
\usepackage[T1]{fontenc}
\usepackage{graphicx}
\usepackage{amsmath}
\usepackage{hyperref}
\usepackage{amsfonts}
\usepackage{footmisc}
\usepackage{booktabs}
\usepackage{multirow}
\usepackage{amssymb}
\usepackage{authblk}
\usepackage{lmodern}
\usepackage{makecell}  
\usepackage{float}
\begin{document}
\bibliographystyle{plain}
\title{VisionLLM-based Multimodal Fusion Network for Glottic Carcinoma Early Detection}
\titlerunning{Multimodal Glottic Carcinoma Network}
%
%
\author{Zhaohui Jin\inst{1}\textsuperscript{\textasteriskcentered} \and
Yi Shuai\inst{2}\textsuperscript{\textasteriskcentered} \and
Yongcheng Li\inst{1} \and
Lingcong Cai\inst{1} \and
Yun Li\inst{2} \and
Huifen Liu\inst{1}\textsuperscript{\dag} \and
Xiaomao Fan\inst{1}\textsuperscript{\dag}}
\authorrunning{F. Author et al.}
%
\institute{Shenzhen Technology University, Shenzhen, China \and
First Affiliated Hospital of Sun Yat-sen University, Guangzhou, China}
\maketitle              
\footnotetext[1]{Zhaohui Jin and Yi Shuai contribute equally.}
\footnotetext[2]{Huifen Liu and Xiaomao Fan are the corresponding authors. Email: liuhuifen@sztu.edu.cn; astrofan2008@gmail.com. }
\begin{abstract}

The early detection of glottic carcinoma is critical for improving patient outcomes, as it enables timely intervention, preserves vocal function, and significantly reduces the risk of tumor progression and metastasis. However, the similarity in morphology between glottic carcinoma and vocal cord dysplasia results in suboptimal detection accuracy. To address this issue, we propose a vision large language model-based (VisionLLM-based) multimodal fusion network for glottic carcinoma detection, known as MMGC-Net. By integrating image and text modalities, multimodal models can capture complementary information, leading to more accurate and robust predictions. In this paper, we collect a private real glottic carcinoma dataset named SYSU1H from the First Affiliated Hospital of Sun Yat-sen University, with 5,799 image-text pairs. We leverage an image encoder and additional Q-Former to extract vision embeddings and the Large Language Model Meta AI (Llama3) to obtain text embeddings. These modalities are then integrated through a laryngeal feature fusion block, enabling a comprehensive integration of image and text features, thereby improving the glottic carcinoma identification performance. Extensive experiments on the SYSU1H dataset demonstrate that MMGC-Net can achieve state-of-the-art performance, which is superior to previous multimodal models. 

\end{abstract}

\section{Introduction}
Glottic carcinoma is a common malignant tumor of the head and neck that can severely affect the patient's voice, swallowing, breathing function, and even overall health \cite{KWON2022}. The early detection of glottic carcinoma heavily relies on the manual interpretation of laryngoscopic results. However, vocal cord dysplasia, a kind of precancerosis is hard to distinguish from glottic carcinoma under laryngoscopy \cite{10.1158/0008-5472.CAN-14-1458}. Moreover, the small size of lesions, coupled with the limitations of specialist expertise may lead to a high misdiagnosis rate \cite{irjala2011pharyngo,azam2022deep,tamagawa2015primary}. Therefore, the development of an automated approach to assist laryngologists in the early detection of glottic carcinoma is of paramount importance. 


Recently, significant progress has been made in deep learning techniques for tackling real-world classification tasks in computer vision and natural language processing. Many researchers have sought to apply these models to the detection of laryngeal cancer, yielding promising outcomes. However, most existing methods only utilize laryngoscopic images as input, including UC-DenseNet \cite{luo2022diagnosis}, MTANet \cite{zhou2022mtanet}, DLGNet \cite{wang2023dlgnet}, RedFormer \cite{cui2023redformer}, and SAM-FNet \cite{wei2024sam}. Although these methods have demonstrated improved performance in laryngeal cancer detection, they mainly neglected the potential benefits of incorporating text modality to further enhance classification accuracy. Therefore, we have curated the SYSU1H dataset from the First Affiliated Hospital of Sun Yat-sen University, which includes both laryngoscopic images and corresponding clinical reports. This comprehensive dataset enables the integration of multimodal information, offering a significant advantage in improving the performance and robustness of classification models.

In this paper, we propose a simple yet effective multimodal fusion framework, termed MMGC-Net, for early detection of glottic carcinoma. Specifically, we introduce a laryngoscopic image encoder \cite{li2023blip} to extract image embeddings from laryngoscopic images and Llama3 \cite{dubey2024llama} to derive text embeddings from clinical reports. Additionally, we design a laryngeal feature fusion block that integrates these two modalities, enabling the model to capture joint features from both the visual data and the corresponding clinical reports. The combination of these modalities allows the model to better exploit the complementary information provided by the images and texts, enhancing its classification capability. Extensive experiments on our collected SYSU1H dataset demonstrate that our proposed MMGC-Net can achieve state-of-the-art results with a significant margin. Overall, our main contributions can be summarized as follows:

\begin{itemize}
    \renewcommand{\labelitemi}{\textbullet}
    \item  We present the SYSU1H dataset, collected from the First Affiliated Hospital of Sun Yat-sen University, which includes both laryngoscopic images and clinical reports, providing a unique resource for laryngeal cancer detection. 
    \item  We employ the pre-trained image encoder and Q-Former to extract well-aligned features from laryngoscopic images, and introduce Llama3, leveraging its advanced natural language processing capabilities, to extract text embeddings from clinical reports. 
    \item We propose MMGC-Net, a novel VisionLLM-based multimodal fusion network for the early detection of glottic carcinoma, which is the first application of multimodal learning to this specific task.
    \item Extensive experiment results conducted on our collected SYSU1H dataset demonstrate that our proposed MMGC-Net can achieve state-of-the-art results with a significant margin. 
\end{itemize}

\section{Methodology}

\begin{figure}[tb]
    \includegraphics[width=1.00\linewidth]{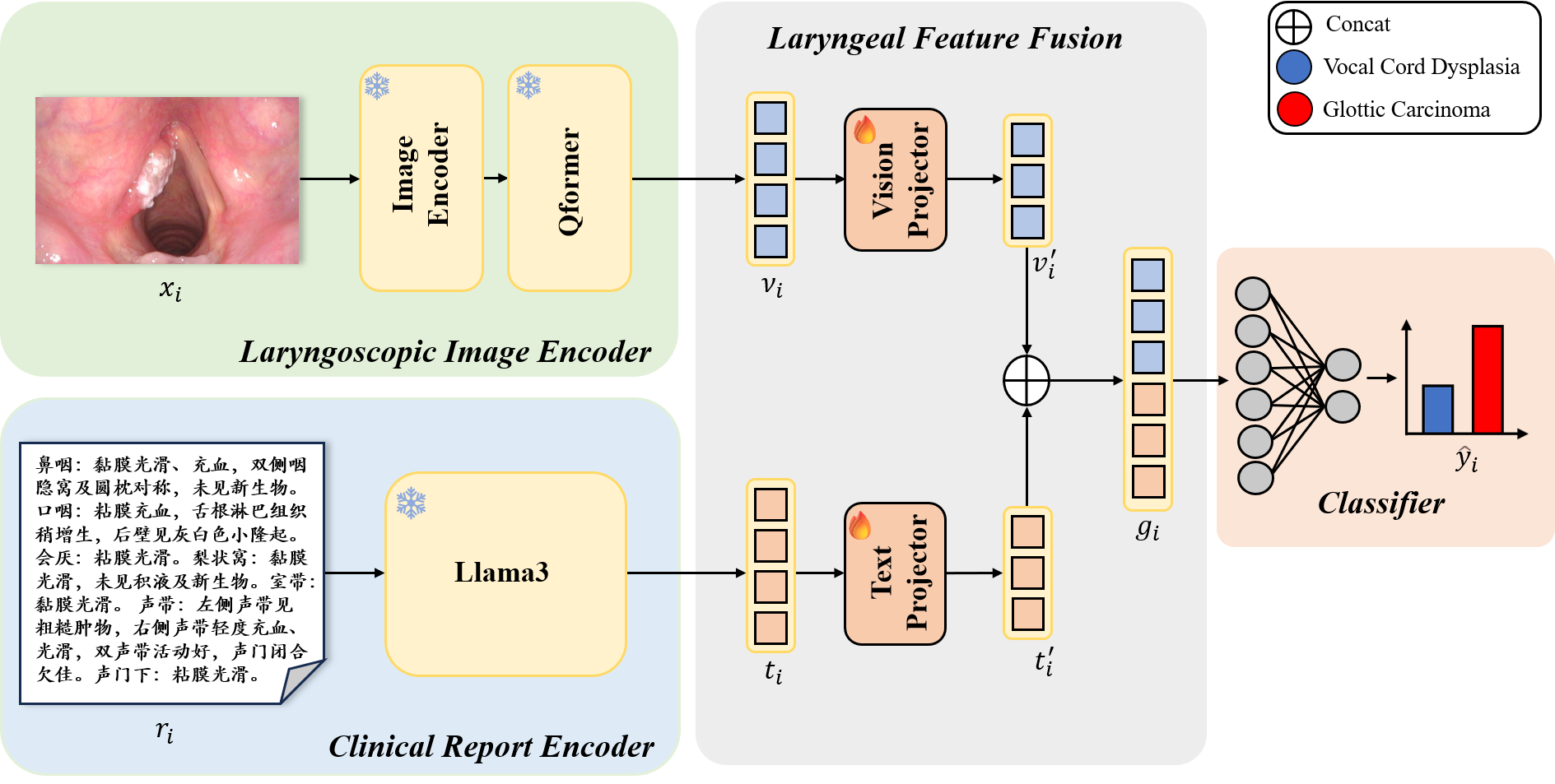}
    \caption{The overall architecture of our proposed MMGC-Net.}
    \label{framework}
\end{figure}


Figure \ref{framework} illustrates the overall architecture of our proposed MMGC-Net, which consists of three main components: laryngoscopic image encoder, clinical report encoder, and laryngeal feature fusion block. The laryngoscopic image encoder extracts embeddings from laryngoscopic images, while the text encoder processes clinical reports to obtain text embeddings. These embeddings are then integrated via the laryngeal feature fusion block. Finally, a fully connected layer is utilized for classification. The detailed information is outlined as follows:

\subsection{Laryngoscopic Image Encoder}
Solely utilizing an image encoder (i.e. ViT\cite{alexey2020image}) to extract image features may result in significant discrepancies between image and text features, potentially resulting in a decline in classification performance. To address this issue, we leverage an additional Q-Former \cite{li2023blip} in our laryngoscopic image encoder to extract the image feature most relevant to the text. Formally, let $f_{enc}(\cdot)$ and $f_{q}(\cdot)$ represent the image encoder and Q-Former respectively. Given a image $x_{i}$, the image feature $v_{i}$ can be obtained by:
\begin{equation}
    v_{i} = f_{q}(f_{enc}(x_{i};\theta_{enc});\theta_{q}),
\end{equation}
\noindent
where $\theta_{enc},\theta_{q}$ are the parameter of image encoder and Q-Former. It is noteworthy that we use the pre-trained weight of BLIP-2 as the parameters for our laryngoscopic image encoder, since a great number of studies have demonstrated the effectiveness of BLIP2 in downstream tasks \cite{lee2023personalizing,nguyen2024improving,zhu2023chatgpt}. It ensures that the extracted image features are well-aligned with the text features, which can mitigate the catastrophic consequences that could arise from modality discrepancies.



\subsection{Clinical Report Encoder}
For the text modality, we employ Llama3, a sophisticated large language model renowned for its robust multilingual and long-text processing capabilities. Compared to other large language models, Llama3 excels in handling Chinese clinical reports, making it particularly well-suited for our application. Formally, let \( f_{Llama3}(\cdot) \) denote the clinical report encoder. Given a clinical report \( r_i \), the encoder can be expressed as:

\begin{equation}
    t_{i} = f_{Llama3}(r_{i}, \theta_{l}),
\end{equation}

\noindent
where \( t_{i} \) represents the text embedding generated by the clinical report encoder, and \( \theta_{l} \) denotes the encoder’s parameters.

\subsection{Laryngeal Feature Fusion}
To fuse the image and text modalities, we introduce a laryngeal feature fusion block. Specifically, we first leverage the vision projector \(f_{vp}(\cdot)\) and text projector \(f_{tp}(\cdot)\), which map the image and text embeddings \(v_i\) and \(t_i\) into a unified representation space, formulated as:
\begin{eqnarray}
    v'_{i}&=&f_{vp}(v_{i};\theta_{vp}),\\
    t'_{i}&=&f_{tp}(t_{i};\theta_{tp}),
\end{eqnarray}
\noindent where \(v'_{i}\) and \(t'_{i}\) represent the transformed image and text features, and \(\theta_{vp}, \theta_{tp}\) denote the respective parameters of \(f_{vp}(\cdot)\) and \(f_{tp}(\cdot)\). Additionally, we apply L2-normalization to both features, yielding \(v''_{i} = \frac{v'_{i}}{\|v'_{i}\|_2}\) and \(t''_{i} = \frac{t'_{i}}{\|t'_{i}\|_2}\). Finally, the normalized features \(v''_{i}\) and \(t''_{i}\) are concatenated to form the vision-language joint feature, denoted as \(g_i=Concat(v''_{i}, t''_{i})\). The vision-language joint feature $g_i$ is then passed through a classifier $f_{fc}$, which can be defined as follows:
\begin{equation}
    \hat{y}_i = f_{fc}(g_i,\theta_{fc})
\end{equation}
\noindent
where $\theta_{fc}$ represents the parameters of the classifier and \(\hat{y}\) is the predicted probability distribution by the model. To optimize the MMGC-Net, this study utilizes the cross-entropy loss function, denoted as $\mathcal{L}_{ce}$, which can be expressed as follows:
\begin{equation}
    \mathcal{L}_{CE} = -\sum_{i=1}^{C} y_i \log(\hat{y}_i)
\end{equation}
\noindent
where \(C\) is the number of classes and \(y\) represents the one-hot encoded ground truth labels.

\section{Experiment}

\subsection{Dataset}
\textbf{SYSU1H:} The dataset, collected from the First Affiliated Hospital of Sun Yat-sen University, comprises 5,799 image-text pairs classified into two categories: vocal cord dysplasia and glottic carcinoma. Each image-text pair contains a laryngoscopic image and a corresponding laryngoscopy report (written by a professional doctor). This dataset is collected and organized for the first time, and it has unique clinical value, which can effectively support the early detection research of glottic carcinoma. The dataset has been split into training, validation, and test sets following an 8:1:1 ratio, providing sufficient data support for this study and ensuring the reliability and representativeness of the experimental results.

\subsection{Implementation Details}

All experiments are carried out on a dedicated server featuring eight NVIDIA A6000 GPUs with a total of 196GB of video memory. The system runs on Ubuntu 20.04.5 LTS, utilizing Pytorch 1.9.1 and Scikit-learn 1.3.1 for the implementation. In this study, we employ the AdamW to optimize MMGC-Net with an initial learning rate of 0.00001, with a warm-up and cosine learning schedules to dynamically control the learning rate. MMGC-Net is trained for 80 epochs.

\subsection{Experimental Results}

\begin{table*}[tb]

\center
\caption{Comparison with the state-of-the-art multimodal machine learning models. The best performance is in \textbf{bold} and the second best is indicated with \underline{underline}.}
\resizebox{\textwidth}{!}{
\begin{tabular}{c c c c c c c}

\hline
\multirow{2}{*}{Methods}&\multicolumn{4}{c}{Overall results} & \multicolumn{2}{c}{Recall for different classes}\\ 
\cmidrule(lr){2-5}\cmidrule(lr){6-7}
& Accuracy(\%) & Precision(\%) & Recall(\%) & \(F_1\) score(\%) & VCD(\%) & GC(\%)\\ 
\hline
\specialrule{0em}{2pt}{0pt}
CLIP\cite{radford2021learning} & \underline{67.24±2.7} & \underline{67.49±2.6} & \underline{66.26±2.7} & \underline{67.57±3.6} & \underline{63.76±7.0} & 70.75±3.8 \\
BLIP\cite{li2022blip} & 58.95±1.9 & 59.30±1.7 & 59.02±1.9 & 59.13±1.1 & 49.31±7.5 & 68.70±4.2\\
VILT\cite{kim2021vilt} & 59.10±0.1 & 59.07±0.1 & 59.08±0.1 & 59.07±0.1 & 49.13±4.9 & 69.01±4.4\\
ALIGN\cite{jia2021scaling} & 62.31±2.0 & 62.51±0.9 & 56.81±3.3 & 61.80±1.9 & 33.13±7.9 & \underline{80.48±8.3}\\
MMGC-Net (Ours) & \textbf{76.10±0.2} & \textbf{76.70±0.2} & \textbf{76.16±0.2} &\textbf{74.41±0.5} & \textbf{68.83±1.4} & \textbf{83.48±1.1}\\
\hline
\end{tabular}
}
\label{results}
\end{table*}

To demonstrate the effectiveness of our proposed MMGC-Net, we compare our method with four state-of-the-art methods, including CLIP \cite{radford2021learning}, BLIP \cite{li2022blip}, VILT \cite{kim2021vilt}, and ALIGN \cite{jia2021scaling}. The average results from five trials of various benchmark models, along with our proposed model, are presented in Table \ref{results}. MMGC-Net demonstrates promising performance, achieving accuracy, precision, recall, and \(F_1\) score of 76.10\%, 76.70\%, 76.16\%, and 74.41\%, respectively. Notably, our model surpasses the second-best model by 8.86\% and 9.90\% in accuracy and recall, respectively. Regarding the single-class recall, MMGC-Net achieves 68.83\% for vocal cord dysplasia and 83.48\% for glottic carcinoma. Compared to other state-of-the-art models, MMGC-Net demonstrates significant enhancements in both accuracy and recall, indicating its superior capability in identifying glottic carcinoma.

\subsection{Ablation Studies}

\begin{table*}[tb]
\center
\caption{Ablation study of MMGC-Net.}
\label{ablation table}
\resizebox{\textwidth}{!}{
\begin{tabular}{c c c c c c c}
\hline
Variants & Image & Report & Accuracy(\%) & Precision(\%) & Recall(\%) & \(F_1\) score(\%)\\ 
\hline

M1 & \checkmark &  & 66.72±2.6 & 67.82±2.5 & 67.44±3.0 & 67.63±2.7 \\
M2 &  & \checkmark & 65.36±0.8 & 71.51±2.4 & 64.56±1.6 & 61.59±1.3 \\
M3 & \checkmark & \checkmark & 76.10±0.2 & 76.70±0.2 & 76.16±0.2 & 74.41±0.5\\
\hline
\end{tabular}
}
\end{table*}
To verify the effectiveness of our method for early detection of glottic laryngeal cancer, we compare variants of our method by using only the image modality or text modality: (1) M1 represents using only the image modality; (2) M2 represents using only the text modality; (3) M3 represents using both image and text multimodal data.

As shown in table \ref{ablation table}, The model variant M3, which integrates both image and text modality, demonstrates the best overall performance across all metrics. Specifically, M3 achieves an accuracy of 76.10\%, precision of 76.70\%, recall of 76.16\%, and an \(F_1\) score of 74.41\%, which outperforms M1 and M2 with a significant margin. These results clearly indicate that using both image and report data significantly enhances the model's performance compared to using either modality alone.

\section{Conclusion}
In this paper, we propose MMGC-Net, a novel model for the early detection of glottic carcinoma. MMGC-Net integrates a laryngoscopic image encoder, a clinical report encoder, and a laryngeal feature fusion block. These two modules efficiently extract embeddings from laryngoscopic images and clinical reports, respectively. Leveraging the laryngeal feature fusion block, MMGC-Net captures intrinsic patterns from these multimodal embeddings. Extensive experiments on our SYSU1H dataset demonstrate that MMGC-Net achieves a new state-of-the-art baseline with a substantial performance improvement.

\section{Acknowledgment}
This work is partially supported by the National Natural Science Foundation of China (62473267), the Basic and Applied Basic Research Project of Guangdong Province (2022B1515130009), the Special subject on Agriculture and Social Development, Key Research and Development Plan in Guangzhou (2023B03J0172), and the Natural Science Foundation of Top Talent of SZTU (GDRC202318).

%
%
%

%




%

\end{document}